\documentclass[letterpaper, 10 pt, conference]{ieeeconf}  

\IEEEoverridecommandlockouts                              

\usepackage{amssymb}  
\usepackage{fancyvrb}
\usepackage{paralist}
\usepackage{color,soul}
\usepackage{tikz}
\usepackage{tabularx}
\usepackage{graphicx}
\graphicspath{{./images/}}
\usepackage{siunitx}
\usepackage{url}

\title{\LARGE \bf
BittyBuzz: A Swarm Robotics Runtime for Tiny Systems
}

\author{Ulrich Dah-Achinanon, Emir Khaled Belhaddad, Guillaume Ricard and Giovanni Beltrame$^{1}$
\thanks{*This work was supported by the Fonds de recherche du Québec – Nature et technologies (FRQNT) under grant 296737 and by the National Research Council Canada (NRC).}
\thanks{$^{1}$Ulrich Dah-Achinanon, Emir Khaled Belhaddad, Guillaume Ricard and Giovanni Beltrame are with the Department of Computer and Software Engineering of
        Polytechnique Montreal, Montreal, QC H3T 1J4, Quebec, Canada
        {\tt\small \{ulrich.dah-achinanon, emir-khaled.belhaddad, guillaume.ricard, giovanni.beltrame\}@polymtl.ca}}%
}

\begin{document}

\maketitle
\thispagestyle{empty}
\pagestyle{empty}

\begin{abstract}

Swarm robotics is an emerging field of research which is increasingly
attracting attention thanks to the advances in robotics and its potential
applications. However, despite the enthusiasm surrounding this area of
research, software development for swarm robotics is still a tedious task. That
fact is partly due to the lack of dedicated solutions, in particular for
low-cost systems to be produced in large numbers and that can have important
resource constraints. To address this issue, we introduce BittyBuzz, a novel
runtime platform: it allows Buzz, a domain-specific language, to run on
microcontrollers while maintaining dynamic memory management. BittyBuzz is
designed to fit a flash memory as small as 32 kB (with usable space for
scripts) and work with as little as 2 kB of RAM. In this work, we introduce the
BittyBuzz implementation, its differences from the original Buzz virtual
machine, and its advantages for swarm robotics systems. We show that BittyBuzz
is successfully integrated with three robotic platforms with minimal memory
footprint and conduct experiments to show computation performance of BittyBuzz.
Results show that BittyBuzz can be effectively used to implement common swarm
behaviors on microcontroller-based systems.

\end{abstract}

\section{Introduction}

Swarm robotics is a research area that studies coordination in groups of
relatively simple robots, leveraging local interactions to generate emergent
global behaviors. It is inspired by societies of insects, where groups can
achieve tasks beyond the capabilities of the
individuals~\cite{navarro2013introduction}. With properties such as
scalability, robustness, autonomy, decentralization~\cite{cheraghi2021past},
swarm robotics allow a wide variety of applications, including nanomedicine,
space exploration, search and rescue missions, and generally tasks in dangerous
areas~\cite{cheraghi2021past}. Despite all these possible applications, most of
the field of swarm robotics is still firmly at the research stage. For reasons
related to cost, time, space and complexity, swarm applications require small
and low-cost robots. Multiple robotic platforms meeting these specifications
are widely used for research (e.g. the Kilobot~\cite{rubenstein2012kilobot}).
Due to their low-cost, such systems come with a limited amount of computing
resources and sometimes make swarm-oriented software solutions developed in
simulation unusable. If programming cooperative behaviours for swarm robotics
has always been a challenging task~\cite{yi2020actor}, doing it for
resource-constrained swarm systems only makes it harder.

In fact, the development process for swarm robotics applications is sometimes a
slow and tedious task requiring a lot of effort from the developer. The lack of
dedicated tools focusing on swarm interactions is to blame for this situation:
it forces developers to ``reinvent the wheel'' when trying to integrate
well-known algorithms to new applications. To solve the problem, some previous
work proposed domain specific languages (DSLs) to raise the level of
abstraction and provide common programming primitives, easing the development
task. For instance, Koord~\cite{ghosh2020koord} is an event-driven language
which uses shared variables for coordination in distributed robotics
applications, with capabilities applicable to swarm robotics.
Protelis~\cite{pianini2015protelis} is also an example of a DSL used for
aggregate programming. In the same spirit, the framework
DRONA~\cite{desai2017drona} proposes the P language for distributed mobile
robots applications development. Pinciroli et al. created
Buzz~\cite{pinciroli2016buzz}, a DSL for heterogeneous robot swarms. Buzz gives
a configurable level of abstraction (swarm or individual robot) to the
developer depending on their needs. It also lays down some important primitives
that are necessary for swarm-oriented programming.

Unfortunately, these DSLs do not tackle the resource limitation problem in
their implementations and may therefore be unusable in many practical
applications. In fact, even the Buzz virtual machine, which in theory only
takes 12~kB of memory, can quickly fill the flash when combined with the native
firmware of certain robots. Furthermore, the RAM consumption during
application execution can easily exceed the one available on some robots (as
little as 2~kB for the Kilobot). Faced with such an issue, we propose a system
offering the same capabilities as Buzz for swarm systems composed of
resource-constrained robots.

In particular, this system should offer, to name a few: swarm-level
abstractions, heterogeneous robot support, communication neighbor operations.
The system should also be modular, allowing the user to reduce resource
consumption by selecting the features to use depending on the implemented
behavior. It should be able to fit systems like the Kilobot with 32~kB of
flash and 2~kB of RAM, and other resource-constrained robotic and swarm
research platforms.

Answering these needs, this article introduces BittyBuzz, an implementation of
the Buzz Virtual Machine for microcontrollers. This implementation roughly uses
the same structure as Buzz and supports 100\% of its code, with very few
limitations when compared to Buzz, to address resource constraints. The main
contributions of this paper are:
\begin{itemize}
\item the development of a swarm-oriented runtime environment capable of
  defining the behavior of heterogeneous swarms of robots with as small as
  32~kB of flash and 2~kB of RAM;
\item the adaptation and optimization of Buzz's capabilities (generality, mixed
  bottom-up/top-down logic, etc.) to resource-constrained platforms;
\item the integration of BittyBuzz with three different robot hardware
  platforms used for research in swarm robotics.
\end{itemize}

BittyBuzz is released as open-source software and can be downloaded from our
repository\footnote{\url{https://github.com/buzz-lang/BittyBuzz}}. We evaluated
BittyBuzz's performance on the Bitcraze
Crazyflie, the K-Team kilobot, and the Zooid robotic platforms. All these
platforms are currently used in research on swarm robotics.

The rest of this paper is organized as follows. In Section~2, we discuss some
of the work related to BittyBuzz, while explaining the differences. Section~3
presents BittyBuzz's features and design principles while comparing them to
Buzz's. Section~4 explains the design choices that were made to overcome the
resource constraints. Section~5 presents an overview of the
BittyBuzz-integrated robotic platforms. In Section~6 we evaluate the VM
performance. We then draw the concluding remarks in Section~7.

\section{Related work}

A variety of previous works discussed the development of dedicated tools for
swarm systems, as well as the creation of optimized frameworks for
resource-constrained devices (see Table~\ref{tab:work_comparison} for a
summary). Micropython~\cite{MicroPyt58:online} is a lightweight and efficient
implementation of the Python 3 programming language. That implementation is
optimized to run on microcontrollers and in constrained environments. It is
composed of a full Python runtime and a subset of the Python
standard library while able to fit a 256 KB code space and run with 16~kB RAM.
Micropython has been proven to be a good option for rapid development of IoT
devices with tested and verified libraries~\cite{gaspar2020development}.
Currently, its main use targets are: device development and testing, sensor
design, monitoring and configuration tools in design of complex applications,
and education purposes~\cite{gaspar2020development}. For instance,~\cite{khamphroo2017integrating}
presents a prototype of an educational mobile robot based on MicroPython. Even
by addressing the resource constraint issue, this language uses more memory
than BittyBuzz and does not offer any swarm primitives or neighbor management
that make BittyBuzz tailored for swarm application development.

Artoo~\cite{ArtooPla45:online} is another development tool for microcontrollers
and a micro-framework designed for robotics. The framework provides a powerful
DSL for robot control and physical computing. It works on Ruby and borrows some
concepts and code from Sinatra~\cite{GitHubsi5:online}. Supporting 15 different
platforms, among which the Bitcraze Crazyflie (one of our targets with
BittyBuzz), Artoo allows developers to create solutions that incorporate
multiple, different hardware devices at the same time. However, Artoo is not
designed for decentralized swarm application design: it requires the user to
connect one or multiple robots to a centralized control computer, which is not
suitable for inherently distributed swarm systems.

The Zephyr project~\cite{ZephyrProj2020Sep} is also worth mentioning as it
unites developers and users in building a small, scalable real-time operating
system (RTOS) optimized for resource-constrained devices on multiple
architectures. The goal of the project is to create an open, collaborative
environment to deliver a RTOS that will answer the changing demand of the
connected devices. The system supports multiple boards and is easily
portable across platforms. It has a different approach to the resource
constraint problem when compared to BittyBuzz, simply offering a configurable
RTOS that can be adapted depending on the user's needs.

Other existing works focus on the development of frameworks and/or DSLs for
swarm applications. For instance, Koord~\cite{ghosh2020koord} is a swarm-oriented
language developed to make platform-independent code portable and 
verifiable. Koord proposes various useful features for coordination in a swarm,
such as shared variables between the robots and state recording. However, Koord
does not focus on resource constraints, which is the main contribution of this
paper.

Another example is DRONA~\cite{desai2017drona}, a framework for building
reliable distributed mobile robotics (DMR) applications. DRONA provides a
state-machine based language (P) for event-driven programming. P is a high-level
language that, once compiled, generates C code that can be directly deployed in
ROS. Protelis~\cite{pianini2015protelis} is another language developed to
provide a practical and universal platform for aggregate programming. Aggregate
programming, unlike the traditional device-centric programming, is concerned with
the behavior of a collection of devices - that can be assimilated to a swarm of
robots. The language is hosted in and integrated with Java. Yi et
al.~\cite{yi2020actor} propose an actor-based programming framework for swarm
robotic systems: they present a bottom-up programming approach based on a new
control unit, the ``actor''. An actor is the virtualization of the capabilities
of a given robot, and it allows developers to design cooperative tasks without
dealing with the intricacies of robotic algorithms and specific robot brands.
Similarly to Koord, these frameworks are not suitable to run on
resource-constrained devices.

Peng et al.~\cite{peng2016emsbot} present EmSBoT, a component-based framework
targeting resource-constrained devices (using 13~kB of flash memory and 5 KB of
RAM). It is built upon $\mu$COS-III with real-time support. Even though it
is network-focused, this framework supports the development of swarm robotics
applications, albeit without the swarm-oriented features of BittyBuzz.
OpenSwarm~\cite{trenkwalder2016openswarm} is an OS designed for severely
computationally constrained robots (using 1~kB RAM and 12~kB of flash). It
enables the developer to design platform-independent solutions, that can easily
be applied to swarm robotics, as showed in their experiments. Both EmSBoT and
OpenSwarm only provide low-level programming and work on a small subset of
platforms, while Buzz and BittyBuzz are based on a virtual machine (VM),
meaning that the Buzz code is device-independent and can run on any system
where the virtual machine is running without need of recompilation.

Overall, although some of those frameworks are lightweight, they either do not
focus on resource-constrained systems or do not meet the key requirements of a
successful programming language for swarm robotics (decentralized control,
spatial computing, neighbor communication, etc.~\cite{pinciroli2016buzz}).
Buzz~\cite{pinciroli2016buzz} is a swarm-specific language that was developed
to address the lack of software development tools in swarm robotics. Buzz
offers a multitude of interesting features for swarm application development,
including but not limited to: heterogeneous support, swarm-level abstractions,
neighbor operations and a consensus system (the ``virtual stigmergy''). The
work presented in this paper is an adaptation of Buzz optimized to fit
resource-constrained systems, such as microcontrollers. BittyBuzz supports the
entirety of the Buzz language, with a runtime that can be adapted to different
memory and computation constraints. We extended the support of BittyBuzz to
three hardware platforms (Bitcraze Crazyflie~\cite{preiss2017crazyswarm},
Kilobot~\cite{rubenstein2014kilobot}, and Zooids~\cite{le2016zooids}) which are
targeted towards resource-constrained robot swarms, but without a common
framework or language. BittyBuzz provides a common language that allows the
user to develop behaviors with minimal knowledge of the specifics of the
hardware.

\begin{table*}[!t]
\caption {Comparison of existing frameworks for robots applications development.} \label{tab:work_comparison} 
\begin{center}
\begin{tabular}{ |c|c|c|c|c|c| } 
\hline
\textbf{System/Framework} & \textbf{Swarm support} & \textbf{Heterogeneous} & \textbf{DSL} & \textbf{Embedded} & \textbf{Support for resource constraints}\\
\hline
MicroPython~\cite{MicroPyt58:online} &  & N/A & N/A & \checkmark & \checkmark \\
\hline
Artoo~\cite{ArtooPla45:online} &  & N/A & N/A &  & \checkmark \\
\hline
Zephyr project~\cite{ZephyrProj2020Sep}  &  & N/A & N/A & \checkmark & \checkmark \\
\hline
DRONA~\cite{desai2017drona}  & \checkmark &  & P & \checkmark &  \\
\hline
Protelis~\cite{pianini2015protelis}  & \checkmark & \checkmark & Protelis & \checkmark &  \\
\hline
Koord~\cite{ghosh2020koord}  & \checkmark & \checkmark & Koord & \checkmark &  \\
\hline
Actor-based framework~\cite{yi2020actor} & \checkmark & \checkmark & (unnamed) & \checkmark &  \\
\hline
Emsbot~\cite{peng2016emsbot} & \checkmark & \checkmark & N/A & \checkmark & \checkmark \\
\hline
OpenSwarm~\cite{trenkwalder2016openswarm} & \checkmark & \checkmark & N/A & \checkmark & \checkmark \\
\hline
 Buzz~\cite{pinciroli2016buzz} & \checkmark & \checkmark & Buzz & \checkmark &  \\
\hline
BittyBuzz & \checkmark & \checkmark & Buzz & \checkmark & \checkmark \\
\hline
\end{tabular}
\end{center}
\end{table*}

\section{BittyBuzz structure}

\begin{figure}[!t]
\centering
\includegraphics[width=\columnwidth]{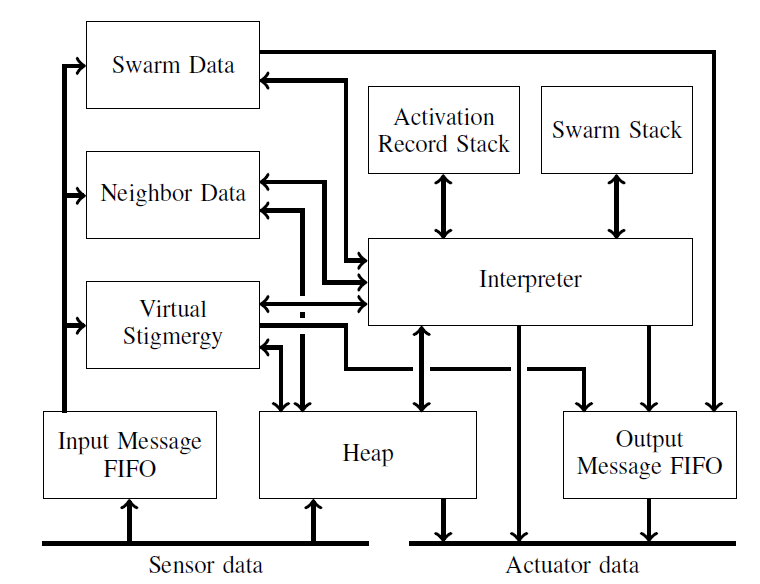}
\caption{BittyBuzz virtual machine structure~\cite{pinciroli2016buzz}}
\label{fig:BBZVM_structure}
\end{figure}

\textbf{BittyBuzz virtual machine.} BittyBuzz, just like Buzz, has a run-time
platform based on a custom virtual machine written in C. The BittyBuzz virtual
machine (BBZVM) structure and operation is the same as Buzz: the virtual
machine operates in disrite time steps, each of which consists in a sequence of
sub-steps: 1) the BBZVM collects the robot's sensor readings, which would
typicalle be stored in memory; 2) the BBZVM collects incoming messages and
updates relevant data structures related to communications (e.g. the
\textit{neighbors} data structure); 3) the BittyBuzz interpreter is called to
execute a Buzz script; 4) and finally the BBZVM outputs actuator signals and
outgoing messages. The BBZVM has a variable size, and can be as small as
17.1~kB (see Section \ref{bbzvm_configuration} for details). Its structure is
presented in Figure~\ref{fig:BBZVM_structure}.

\textbf{Swarm-level abstraction.} A team of robots is considered as a swarm.
Buzz allows the developer to handle robot swarms as a first-class language
object through the \textit{swarm} primity type. This capability is preserved in
BittyBuzz. That way, swarms can be easily created and robots can join or leave
a swarm, optionally based on a specific condition.

New swarms can also be created as a result of an intersection, union or
difference between existing swarms. All the robots in a swarm can then be
assigned a shared task~\cite{pinciroli2016buzz}.

\textbf{Heterogeneous robots support.} Buzz is an extension language and was
explicitly designed to support heterogeneous robot swarms. Hence, it natively
only handles the logic parts of the robotic system (common to all robotic
systems). That allows the developer to extend Buzz by adding new robot-specific
commands (see Section~\ref{sec:bbz_ext} for BittyBuzz
extensions)~\cite{pinciroli2016buzz}.

\textbf{Situated communication.} All the robots running Buzz are assumed to
have a device capable of broadcasting and receiving messages within a limited
range provided direct, unobstructed line-of-sight to and from neighbors. That
mechanism gives each robot the ability to inform its neighbors of its position
and makes each robot aware of its neighbors' positions. When a device receives
a message, it detects the relative distance and angle of the sender, making it
easier to use the information in a swarm application (for instance for obstacle
avoidance). Situated communication is frequently used for different algorithms
in swarm robotics~\cite{pinciroli2016buzz}.

\textbf{Virtual Stigmergy.} Buzz implements the Virtual Stigmergy (VS), a
conflict-free replicated data structure that is used to provide consensus on a
set of key-value pairs across the entire swarm~\cite{pinciroli2016tuple}. The VS
is essentially seen by the programmer as a shared table: each \textit{put}
operation by a member of the swarm triggers an automatic update in the others,
with mechanisms to avoid conflicts. The VS is available in BittyBuzz with its
complete features, and it is completely transparent to the user as a key-value
data store through the \textit{put} and \textit{get} methods.

\textbf{Neighbor operations.} Buzz and BittyBuzz provide a \textit{neighbors}
data structure that allows data collection and processing from neighboring
robots. This type of neighbor operation is the foundation of spatial
computing~\cite{zambonelli2004spatial} and widely used in swarm robotics. The
neighbors structure is a dictionary indexed by robot id containing the robot
distance as well as the azimuth and elevation angles of each neighboring robot,
as detected by the BBZVM, and updated at each step through \textit{situated
communication}. It supports three spatial operations: iteration, transformation
and reduction, with functions such as \textit{map(), foreach(), reduce()} and
\textit{filter()}. The only difference with Buzz is that the functions
\textit{kin()} and \textit{nonkin()} that allow the programmer to filter
neighbors based on swarm membership are not implemented in BittyBuzz to reduce
its memory footprint.

The \textit{neighbors} data structure also allows direct communication with the
robots in the neighborhood, by the means of a \textit{broadcast} operation. The
neighbors that want to receive messages on a given topic can subscribe to it
with the \textit{listen} function. A robot can then also ignore the
messages related to a subscribed topic by using the \textit{ignore} function.

Thanks to these features, BittyBuzz lets developers work at different levels of
abstraction according to their needs, develop intuitive code for heterogeneous
swarms and easily access neighborhood information. Thus, one can focus on the
swarm behavior as a whole with a top-down approach or use the robot-wise
operations in a bottom-up fashion. Those are all essential features
for a swarm-oriented programming language, and are all implemented in
BittyBuzz.

\subsection{BittyBuzz virtual machine configuration}
\label{bbzvm_configuration}
Reducing the BBZVM's size to fit smaller microcontrollers needed some
adaptations to the original Buzz virtual machine. Besides the two unimplemented
functions mentioned above, we provide some parameters to modulate the BBZVM's
size and memory consumption depending on the application. The configurable
attributes range from heap and stack size definition to optional feature
(de)activation.

The swarm structure as well as its operations can be disabled if needed. In this
case, the swarm level abstraction will no longer be exploitable, and all the
instructions will be robot-wise. The neighbor operations can also be disabled by
the user, removing the situated communication. Similarly, the virtual stigmergy
structure can be deactivated.

BittyBuzz also has a memory usage reduction mode that allows the developer to
drastically reduce RAM consumption at the expense of the flash.

All these configurations are selected at compilation time by the user, depending
on the hardware and application.

\subsection{BittyBuzz extension}\label{sec:bbz_ext}
The ability to extend BittyBuzz (and Buzz) is a particularly useful feature when
using the language for a new robotic platform. For instance, the \textit{goto()}
function's implementation and number of parameters is highly platform dependent
and can be redefined for all the new robots by using external C closures.
Defining such closures in BittyBuzz can be done in a very similar way to Buzz.
In BittyBuzz, to use a new function in the Buzz script is as simple as writing
a C function. The created function can then be registered in the BBZVM with a
single call to \textit{bbzvm\_function\_register(fnameid, funp)}.

\section{BittyBuzz: overcoming resource constraints}

Different strategies were used to make BittyBuzz efficient in terms of resource
consumption. This section will present the dynamic memory management mechanisms
implemented, as well as other important optimization points.

\subsection{Dynamic memory management}

\subsubsection{Pre-allocated heap}

\begin{figure}[!t]
  \centering \includegraphics[width=\columnwidth]{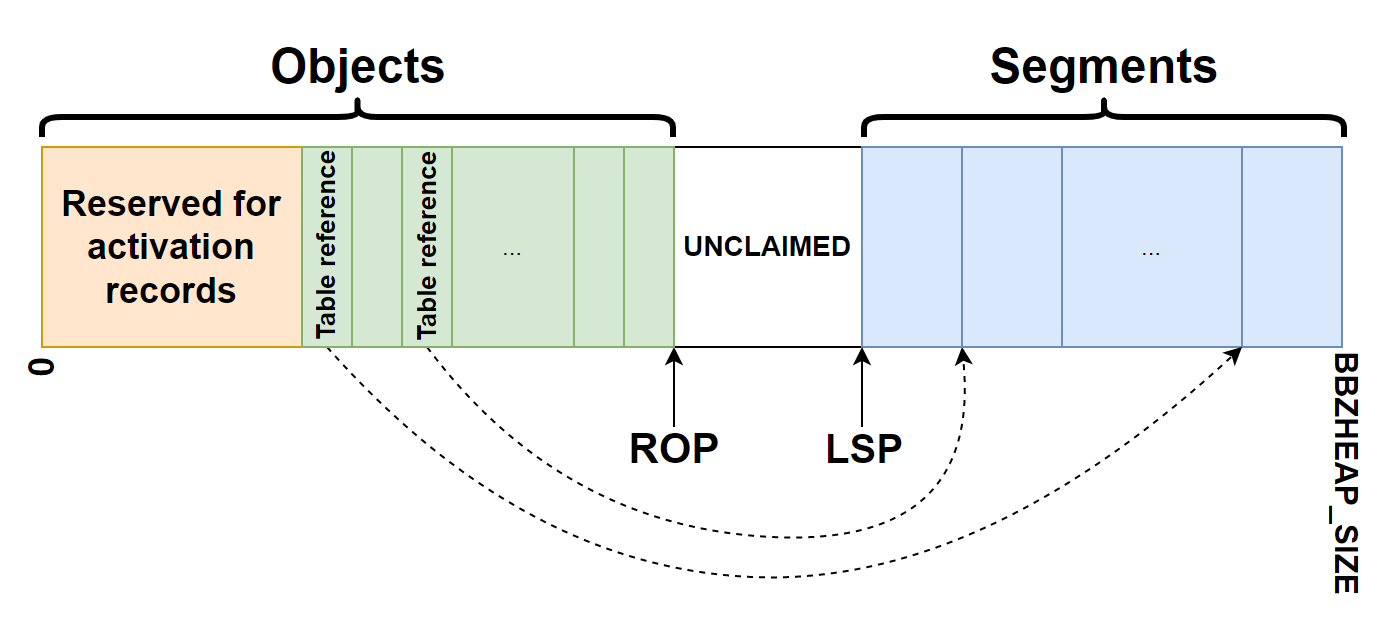}
  \caption{Illustration of BittyBuzz's heap: the object section containing the
    activation records section and the allocated objects (non-structured types
    and table references) in light green, the unclaimed section in white, the
    segment section in light blue, and the two pointers ROP and LSP.}
\label{fig:heap_representation}
\end{figure}

BittyBuzz has a pre-allocated heap whose size is specified by the developer.
The heap is represented by a static buffer with three sections: the object
section (also containing the activation records of closure calls), the segment
section and the unclaimed section. Two pointers are used to access the heap:
the rightmost object pointer (ROP) and the leftmost table segment pointer (LSP)
as shown in Figure~\ref{fig:heap_representation}. Each object has two bytes
containing the payload information and an additional metadata byte.

Regarding the storage in the heap, non-structured types such as nil, int, float
and string are written from left to right with ROP always pointing to the last
added object. For structured types (tables), an object is stored in the
objects' section, referring to a data segment stored from right to left in the
segments section. Each data segment has a user-defined number of key-value
pairs and two metadata bytes containing information about the segment validity
and a pointer to the next segment, if any. To save space in the case of arrays,
the key-value pairs are seamlessly replaced by the values in the case where the
keys are integers, in a way that resembles the table management in
Lua~\cite{jucs05pd57:online}.

\subsubsection{Garbage collector}

BittyBuzz has a simple and effective garbage collector algorithm which is
frequently called during the script execution, specifically before the
execution of instructions. Each BittyBuzz heap entry has an attribute
containing metadata, including some garbage collector bits. The algorithm
starts by unmarking all the segments and objects. Then, it marks
back the permanent objects that should never be garbage-collected, and the
valid variables on the BittyBuzz stack. Finally, all the remaining unmarked
objects are invalidated, and the heap pointers are re-positioned.

\subsection{Optimizations and limitations}
\subsubsection{Ring-buffers}

To reduce the time complexity for message queue management, BittyBuzz
implements its own ring buffer. Ring buffers are used in several algorithms and
are desirable in data stream use cases. They are buffers that are implemented
to seamlessly loop on themselves, allowing constant time \textit{push} and
\textit{pop} operations for queues. Previous works such
as~\cite{feldman2015wait} propose a new implementation for multi-thread
modifications of the ring buffer: we use a similar but simpler implementation
running in a single thread. Each buffer is stored as an element size, a buffer
capacity, a start index and an end index. With this information, insert, pop,
and index access are done in constant time. We use ring buffers in BittyBuzz to
store message queues, the communication payload buffer, and the
\textit{neighbors} data structure.

\subsubsection{Translated bytecode}
The available flash memory is also an important factor in the resource
consumption management. To reduce the storage space, the original Buzz object
(.bo file) is converted into a smaller BittyBuzz object (.bbo file). The
optimizations mainly include the conversion from Buzz integers on 32 bits to
BittyBuzz integers, which are on 16 bits. Floats are also converted to
\textit{bbzfloat} that only use 16 bits. All the non-instruction related strings
are also stripped from the new object file to minimize file size.

\subsubsection{Optimized loops}
As discussed earlier, BittyBuzz uses a ring
buffer to store neighbors' information. In fact, instead of using table
segments on the heap, the implementation uses a static table of user-defined
size for the neighbors' data. We optimize the looping through this data
structure with a \textit{foreach} operation: for each neighbor in the ring
buffer, a data table is created to execute the needed instructions and then
garbage-collected at the end of the iteration. The loop therefore uses constant
instead of linear space on the heap.

\subsubsection{Swarm lists}
In BittyBuzz, a robot can be a member of a maximum of 8 swarms, numbered from 0
to 7. To keep track of which swarms a robot is a member of, we use a swarm list
that is simply an 8-bit variable where the i-th bit represents whether a robot
is a member of the i-th swarm (for example, if the bit 0 is 1 that means the
robot is a member of the swarm 0). This list allows one to easily determine all
the swarms of a given robot. When a robot is added to a swarm, BittyBuzz
foresees to add the robot's swarm list to the outgoing message queue so that its
neighbors can have the information and add it to their swarm table. 

\subsubsection{Virtual stigmergy}
BittyBuzz, just like Buzz, has a virtual stigmergy system, although BittyBuzz's
has some limitations with respect to the original version. For instance,
BittyBuzz only has a unique instance of the stimergy, whereas one could create
an arbitrary number of stigmergies in Buzz. In addition, the current
implementation only takes strings as topics (keys) for the stigmergy.

\subsubsection{String manager}\label{sec:str_man}
Strings are a data type used in many programming languages to represent a
sequence of characters. In BittyBuzz, in addition to variables used in the Buzz
script, strings are used to identify global symbols and function names. In
conventional implementations of strings, the data type is represented as an
array of bytes (or words)~\cite{busbee2018string}. However, such an
implementation is memory consuming and needs an optimization to be effectively
used in a resource constrained system. BittyBuzz represents strings as
integers, by assigning a unique ID to each of the strings used in a given
script, while the string content is placed in the flash RAM. This
implementation reduces the memory footprint for strings as they always occupy 2
bytes instead of an array of unknown bytes. Some string IDs are necessary for
BittyBuzz to operate correctly, and they are natively declared in the VM (for
example the \textit{stigmergy} keyword). As needed, the developer can create
new string IDs for user-defined closures, using the macro
\textit{BBZSTRING\_ID} (see~\ref{sec:bbz_ext}).

\subsubsection{Neighbors communication} As mentioned earlier, BittyBuzz
provides a publish-subscribe like mechanism, to allow robots to communicate. To
send a message, a program calls the function
\textit{\textbf{neighbors}.broadcast(topic, value)}. Because of the extremely
limited payload size of the targeted robotic platforms, BittyBuzz does not
support sending tables as value. Also, since strings in BittyBuzz are
represented by unique IDs (see~\ref{sec:str_man}) created during the execution,
the code on the robots must be the same if the developer is trying to implement
a communication-based behaviour. Specifically, the order and the number of
strings literals must be the same. This final constraint is not limiting, as
using the same script for all robots is the standard way Buzz scripts are
operating.

\begin{figure}[!t]
\includegraphics[width=\linewidth]{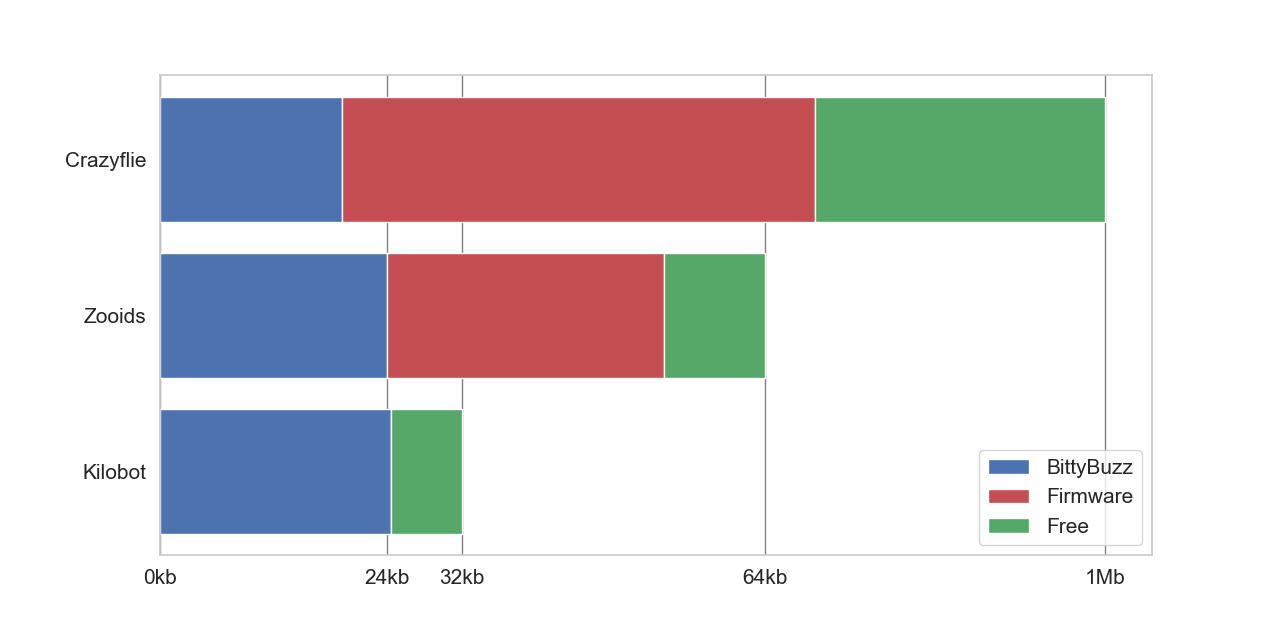}
\caption{Sizes of the BBZVM, the robot native firmware, and the available space
  for user bytecode for the three integrated robotic platforms: Bitcraze Crazyflie,
  K-Team Kilobot and Zooids}
\label{fig:mem_footprint}
\end{figure}

\section{Robotic platforms}

Figure~\ref{fig:mem_footprint} presents the flash memory usage of the BittyBuzz
virtual machine and full-feature framework on the three following robotic
platforms used in different swarm robotic projects. In all instances the memory
footprint is under 24~kB with some variability between platforms based on
available compiler optimizations (eg. floating point hardware support).

\begin{figure}[!t]
\centering
\includegraphics[width=\columnwidth]{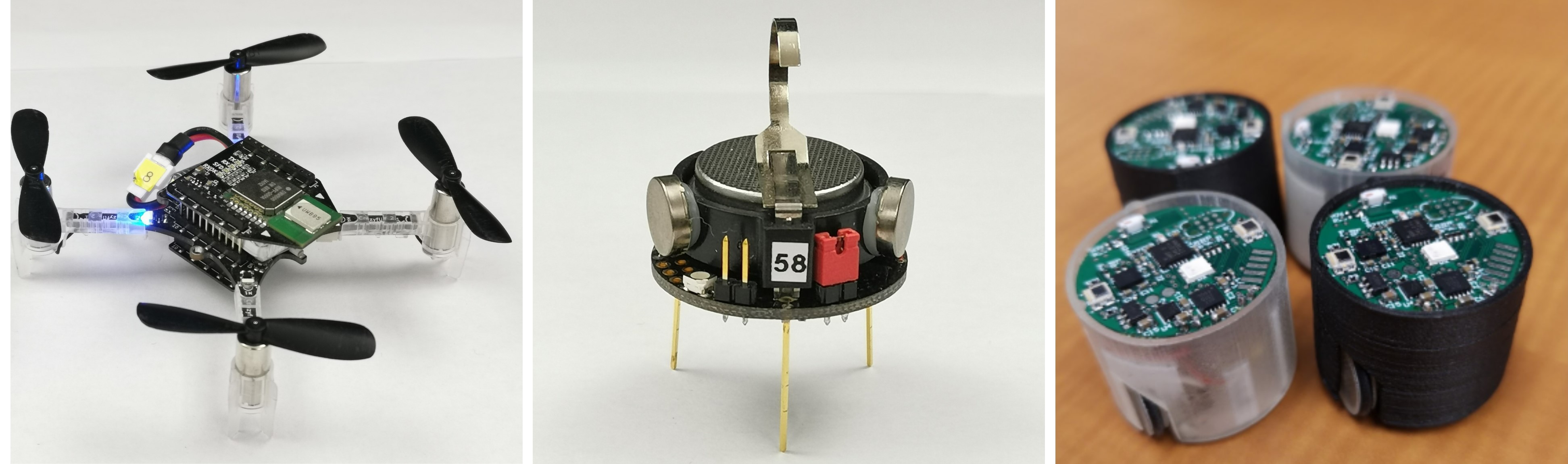}
\caption{Robotic platforms integrated with BittyBuzz. From left to right: Bitcraze Crazyflie, K-Team Kilobot, and Zooids}
\label{fig:robots}
\end{figure}

The \textbf{Kilobot} is an open-source low-cost small robot (33~mm diameter and
34~mm height) designed to ease the testing of algorithms for a large swarm
(hundreds or thousands of robots), which can normally be time- and
money-consuming. Each Kilobot can be made with US\$14 worth of parts and can be
assembled in a relatively short time~\cite{rubenstein2014kilobot}. Kilobots
provide a programmable controller, basic locomotion and local
communication~\cite{KilobotsSelfOrganizingSystemsResea}. They use an ATmega
328p processor (8 bit@8MHz), a 32~kB flash, 2~kB SRAM, 1~kB EEPROM with a
rechargeable battery, and they can be programmed in
C~\cite{KilobotKTeamCorporation}.

The \textbf{Bitcraze Crazyflie}~\cite{giernacki2017crazyflie} is an open-source experimental
flying development platform used for research and education in robotics. This
relatively low-cost, easily expandable and upgradeable quadrotor is 92x92x29mm
and weighs 27g. It can easily be assembled without any soldering and supports
several expansion decks. Crazyflies are equipped with a STM32F405 MCU Cortex-M4
@ 168MHz, a 192~kB SRAM and a 1~MB flash for the main application. Finally,
they have an 8~kB EEPROM~\cite{Crazyflie21BitcrazeStore}.

\textbf{Zooids}~\cite{le2016zooids} are small open-source and open-hardware robots,
designed for a new class of human-computer interfaces for tabletop swarm
interfaces. They each weigh 12~g for 26 mm in diameter and 21 mm in height. 
They contain a 48~MHz ARM microcontroller (STM32F051C8) equipped with a 64~kB
flash memory and 8~kB SRAM.

\section{Performance}

We measured the overhead of the BittyBuzz runtime platform running on the most
modest hardware platform we had targeted -- the Zooids -- to derive maximum
utilization constraints and compare performance metrics with the original Buzz
virtal machine implementation. We consider the total time taken for underlying
virtual machine operations, message sharing etc. (\textit{overhead}), which
gives us the allocated slot for user-defined behavior; and maximum execution
speed (in VM instructions per second) to determine the amount of work that can
be done in that time frame. We use a 10\,Hz update frequency for reference.

Buzz measurements were performed on the Khepera IV~\cite{kteam2016kheperaiv}, a
wheeled robot running a complete operating system on top of a 800 MHz Cortex-A8
processor. Comparing BittyBuzz with this more complex platform lets us assess
the further benefits of using the Buzz programming language for swarm software
development on embedded systems.

Table \ref{tab:overhead} displays the differences in VM overhead for the two
hardware platforms. We experience a slightly higher delay on the Khepera
despite the faster CPU due to interaction with the operating system eg. network and
input/output buffering. Furthermore, Table \ref{tab:maxinstr} details the Buzz
virtual machine instruction budget for 100\,ms of execution in a timestep, thus
targeting a 10 Hz timestep frequency. The upper bound on the constrained Zooid
hardware occupies 459 instructions (ie. at most 918 bytes) amounting to 13\% of
the remaining space (see Figure \ref{fig:mem_footprint}). The instruction
budget represents the amount of code that can be executed in a timestep, thus
the remaining 87\% can be used to store additional behaviors to switch to and
from during execution.

\begin{table}[h]
  \begin{center}
  \renewcommand*{\arraystretch}{1.66}
  \caption{Virtual Machine overhead}
  \begin{tabularx}{\columnwidth}{|p{0.29\columnwidth}|>{\centering}p{0.30\columnwidth}|>{\raggedleft\arraybackslash}p{0.257\columnwidth}|}
    \hline
    \textbf{VM Implementation} & \textbf{Hardware Platform} & \textbf{VM Overhead} \\
    \hline
    Buzz~\cite{pinciroli2016buzz} & Khepera IV & 4.92\,ms \\
    \hline
    BittyBuzz & Zooids & 4.45\,ms \\
    \hline
  \end{tabularx}
  \label{tab:overhead}
  \end{center}
\end{table}

\begin{table}[h]
  \begin{center}
  \renewcommand*{\arraystretch}{1.66}
  \caption{Maximum Buzz instructions for a ten Hz timestep}
  \begin{tabular*}{\columnwidth}{@{\extracolsep{\fill}}|l|c|r|}
    \hline
    \textbf{VM Implementation} & \textbf{Hardware Platform} & \textbf{Max instructions} \\
    \hline
    Buzz~\cite{pinciroli2016buzz} & Khepera IV & 156~000 \\
    \hline
    BittyBuzz & Zooids & 459 \\
    \hline
  \end{tabular*}
  \label{tab:maxinstr}
  \end{center}
\end{table}

Moreover, we carried out significant experiments and swarm tasks with Zooids in
the context of hierarchical swarms~\cite{varadharajan2022hierarchical}. where a group of
guide robots herd a lighter worker swarm and have it move across the
environment, thus demonstrating the capabilities of our embedded framework.

\section{Conclusion}

We presented BittyBuzz, a novel virtual machine for the Buzz language
designed for resource-constrained microcontrollers. The contributions of this work include:
\begin{inparaenum}
\item the development of a virtual machine for a dynamically typed language with
  dynamic memory management with as little as 32~kB of flash and 2~kB of RAM; 
\item the full inclusion of Buzz's programming model (including neighbor
  queries, consensus, etc.) to a resource-constrained platform; 
\item the integration of BittyBuzz with three different robotic
platforms used for research in swarm robotics.
\end{inparaenum}
In addition to these contributions, we believe that resource-constrained
robotic platforms have an important part to play in the future of the internet
of everything. A domain-specific dynamic language such as BittyBuzz will ease
the development and prototyping of swarm applications. Our experiments show
that BittyBuzz can be used for established swarm behaviors and that the
communication model used by the virtual machine has sufficient performance even
on severely constrained devices, enabling user behaviors to perform smoothly
without noticeable delays. We show behaviors like exploration, bidding, and
consensus building on various robotic platforms in the multimedia attachment,
as well as a complex particle swarm experiment~\cite{varadharajan2022hierarchical}.

\bibliographystyle{IEEEtran}

\bibliography{IEEEabrv,root}

\end{document}